\documentclass[nohyperref]{article}

\usepackage{microtype}
\usepackage{graphicx}
\usepackage{subfigure}
\usepackage{booktabs} 

\usepackage{hyperref}
\usepackage[noend]{algorithmic}

\newcommand{\com}[1]{{\color{black!35}//\,#1}}

\usepackage[accepted]{icml2023}

\usepackage{amsmath}
\usepackage{amssymb}
\usepackage{mathtools}
\usepackage{amsthm}
\usepackage{multirow}
\usepackage{float}
\usepackage{wrapfig}
\newfloat{algorithm}{t}{lop}

\usepackage{xcolor}
\definecolor{human-blue}{rgb}{0.619,0.722,0.827}
\definecolor{machine-orange}{rgb}{0.929,0.765,0.584}

\usepackage[capitalize,noabbrev]{cleveref}

\newcommand{\name}{DetectGPT}
\newcommand{\sm}{p_\theta}

\newcommand{\pd}{\mathbf{d}\left(x,\sm,q\right)}
\newcommand{\pdh}{\mathbf{\hat d}}
\newcommand{\probfull}{machine-generated text detection}
\newcommand{\probfulltitle}{Machine-Generated Text Detection}
\newcommand{\papertitle}{{\name}: Zero-Shot {\probfulltitle}\\ using Probability Curvature}
\newcommand{\papertitleoneline}{Zero-Shot {\probfulltitle} using Probability Curvature}

\theoremstyle{plain}

\theoremstyle{definition}

\theoremstyle{remark}

\usepackage[textsize=tiny]{todonotes}

\icmltitlerunning{\papertitleoneline}

\begin{document}

\twocolumn[
\icmltitle{\papertitle}



\icmlsetsymbol{equal}{*}

\begin{icmlauthorlist}
\icmlauthor{Eric Mitchell}{stan}
\icmlauthor{Yoonho Lee}{stan}
\icmlauthor{Alexander Khazatsky}{stan}
\icmlauthor{Christopher D. Manning}{stan}
\icmlauthor{Chelsea Finn}{stan}
\end{icmlauthorlist}

\icmlaffiliation{stan}{Stanford University}

\icmlcorrespondingauthor{Eric Mitchell}{eric.mitchell@cs.stanford.edu}

\icmlkeywords{chatgpt, detection, zero-shot, text}

\vskip 0.3in
]



\printAffiliationsAndNotice{}  

\begin{abstract}
The increasing fluency and widespread usage of large language models (LLMs) highlight the desirability of corresponding tools aiding detection of LLM-generated text. In this paper, we identify a property of the structure of an LLM's probability function that is useful for such detection. Specifically, we demonstrate that text sampled from an LLM tends to occupy negative curvature regions of the model's log probability function. Leveraging this observation, we then define a new curvature-based criterion for judging if a passage is generated from a given LLM\@. This approach, which we call {\name}, does not require training a separate classifier, collecting a dataset of real or generated passages, or explicitly watermarking generated text. It uses only log probabilities computed by the model of interest and random perturbations of the passage from another generic pre-trained language model (e.g., T5). We find {\name} is more discriminative than existing zero-shot methods for model sample detection, notably improving detection of fake news articles generated by 20B parameter GPT-NeoX from 0.81 AUROC for the strongest zero-shot baseline to 0.95 AUROC for {\name}. See \href{https://ericmitchell.ai/detectgpt}{\texttt{ericmitchell.ai/detectgpt}} for code, data, and other project information.
\end{abstract}

\section{Introduction}
Large language models (LLMs) have proven able to generate remarkably fluent responses to a wide variety of user queries. Models such as GPT-3 \citep{gpt3}, PaLM \citep{palm}, and ChatGPT \citep{chatgpt} can convincingly answer complex questions about science, mathematics, historical and current events, and social trends. While recent work has found that cogent-sounding LLM-generated responses are often simply wrong \citep{lin-etal-2022-truthfulqa}, the articulate nature of such generated text may still make LLMs attractive for replacing human labor in some contexts, notably student essay writing and journalism. At least one major news source has released AI-written content with limited human review, leading to substantial factual errors in some articles \citep{cnet}. Such applications of LLMs are problematic for a variety of reasons, making fair student assessment difficult, impairing student learning, and proliferating convincing-but-inaccurate news articles. 
Unfortunately, humans perform only slightly better than chance when classifying machine-generated vs human-written text \citep{gehrmann-etal-2019-gltr}, leading researchers to consider automated detection methods that may identify signals difficult for humans to recognize. Such methods might give teachers and news-readers more confidence in the human origin of the text that they consume.

\begin{figure}
    \centering
    \includegraphics[width=\columnwidth]{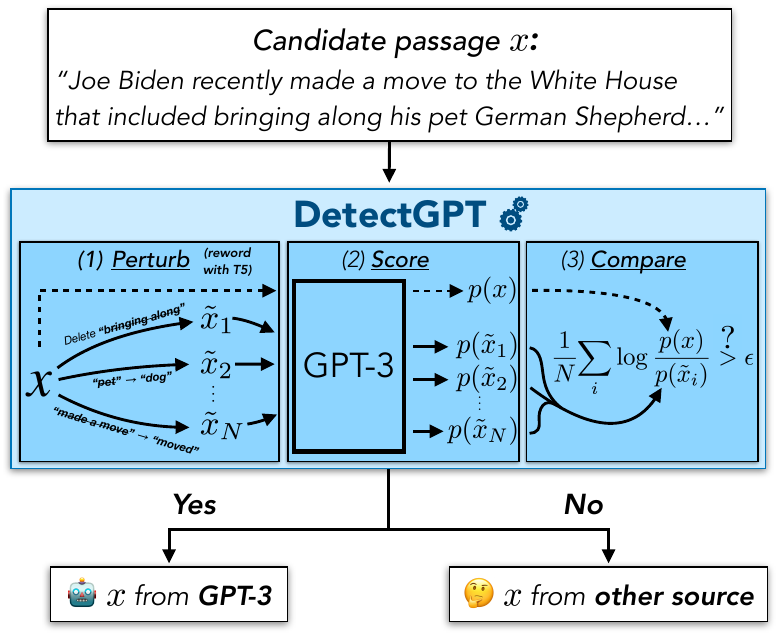}
    \vspace{-6mm}
    \caption{We aim to determine whether a piece of text was generated by a particular LLM $p$, such as GPT-3. To classify a candidate passage $x$, {\name} first generates minor \textbf{perturbations} of the passage $\tilde x_i$ using a generic pre-trained model such as T5. Then {\name} \textbf{compares} the log probability under $p$ of the original sample $x$ with each perturbed sample $\tilde x_i$. If the average log ratio is high, the sample is likely from the source model.}
    \vspace{-4mm}
    \label{fig:fig1}
\end{figure}

As in prior work \citep{Jawahar2020AutomaticDO}, we study the {\probfull} problem as a binary classification problem. Specifically, we aim to classify whether a \textit{candidate passage} was generated by a particular \textit{source model}. While several works have investigated methods for training a second deep network to detect machine-generated text, such an approach has several shortcomings, including a tendency to overfit to the topics it was trained on as well as the need to train a new model for each new source model that is released. We therefore consider the \textit{zero-shot} version of {\probfull}, where we use the source model itself, without fine-tuning or adaptation of any kind, to detect its own samples. The most common method for zero-shot {\probfull} is evaluating the average per-token log probability of the generated text and thresholding \citep{release-strategies,gehrmann-etal-2019-gltr,ippolito-etal-2020-automatic}. However, this zeroth-order approach to detection ignores the local structure of the learned probability function around a candidate passage, which we find contains useful information about the source of a passage.

\begin{figure}
    \centering
    \includegraphics[width=\columnwidth]{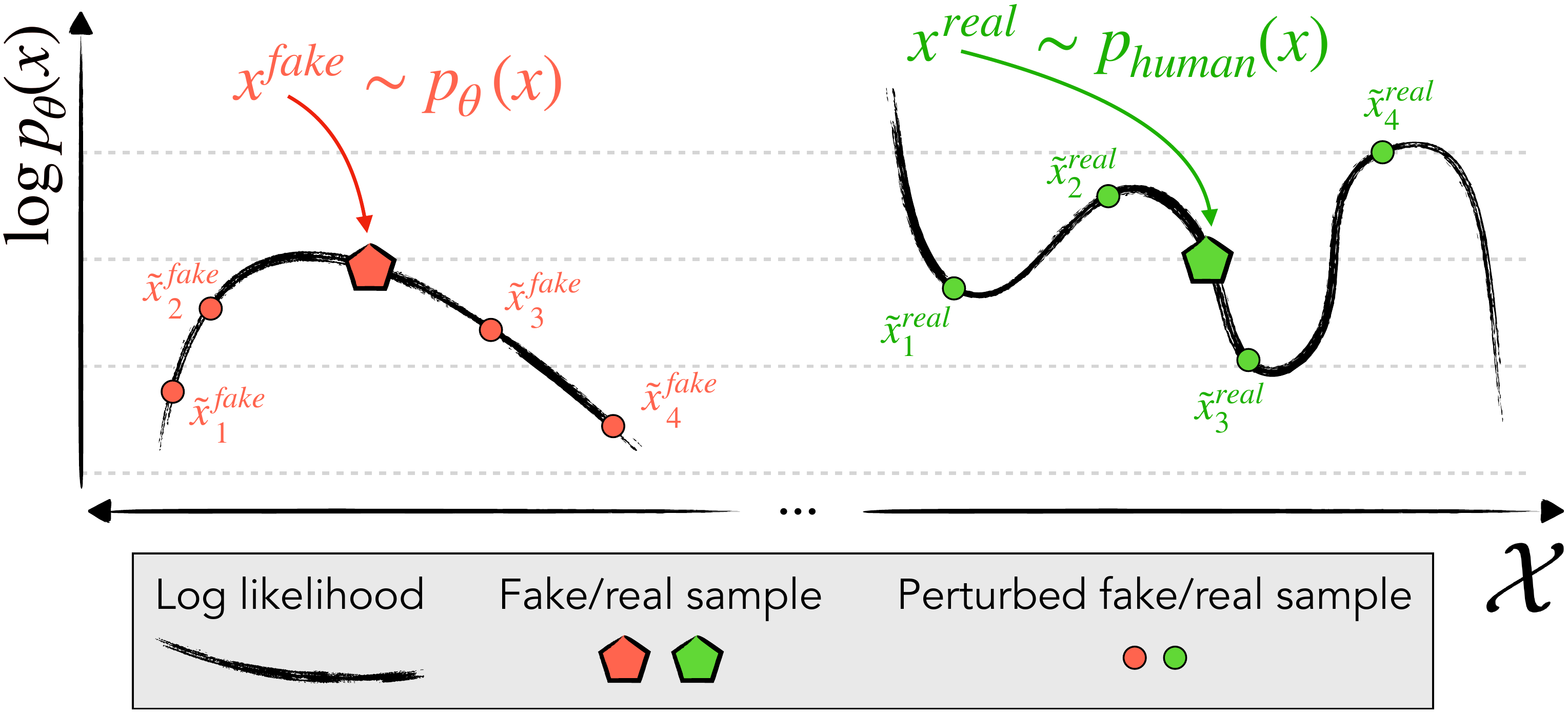}
    \vspace{-5mm}
    \caption{We identify and exploit the tendency of machine-generated passages $x\sim\sm(\cdot)$ \textbf{(left)} to lie in negative curvature regions of $\log p(x)$, where nearby samples have lower model log probability on average. In contrast, human-written text $x\sim p_{real}(\cdot)$ \textbf{(right)} tends not to occupy regions with clear negative log probability curvature; nearby samples may have higher or lower log probability.}
    \vspace{-4mm}
    \label{fig:local-structure}
\end{figure}

This paper poses a simple hypothesis: minor rewrites of \textit{model-generated} text tend to have lower log probability under the model than the original sample, while minor rewrites of \textit{human-written} text may have higher or lower log probability than the original sample. In other words, unlike human-written text, model-generated text tends to lie in areas where the log probability function has negative curvature (for example, near local maxima of the log probability). We empirically verify this hypothesis, and find that it holds true across a diverse body of LLMs, even when the minor rewrites, or \textit{perturbations}, come from alternative language models. We leverage this observation to build {\name}, a zero-shot method for automated {\probfull}. To test if a passage came from a source model $\sm$, {\name} compares the log probability of the candidate passage under $\sm$ with the average log probability of several perturbations of the passage under $\sm$ (generated with, e.g., T5; \citet{raffel2020t5}). If the perturbed passages tend to have lower average log probability than the original by some margin, the candidate passage is likely to have come from $\sm$. See Figure~\ref{fig:fig1} for an overview of the problem and {\name}. See Figure~\ref{fig:local-structure} for an illustration of the underlying hypothesis and Figure~\ref{fig:perturbation-discrepancy} for empirical evaluation of the hypothesis. Our experiments find that {\name} is more accurate than existing zero-shot methods for detecting machine-generated text, improving over the strongest zero-shot baseline by over 0.1 AUROC for multiple source models when detecting machine-generated news articles.

\textbf{Contributions.} Our main contributions are: (a) the identification and empirical validation of the hypothesis that the curvature of a model's log probability function tends to be significantly more negative at model samples than for human text, and (b) {\name}, a practical algorithm inspired by this hypothesis that approximates the trace of the log probability function's Hessian to detect a model's samples.

\section{Related Work}
Increasingly large LLMs \citep{gpt2,gpt3,palm,chatgpt,opt} have led to dramatically improved performance on many language-related benchmarks and the ability to generate convincing and on-topic text. GROVER \citep{Zellers2019DefendingAN} was the first LLM trained specifically for generating plausible news articles. Human evaluators found GROVER-generated propaganda at least as trustworthy as human-written propaganda,
motivating the authors to study GROVER's ability to detect its own generations by fine-tuning a detector on top of its features; they found GROVER better able to detect GROVER-generated text than other pre-trained models. However, \citet{bakhtin2019learning,uchendu-etal-2020-authorship} note that models trained explicitly to detect machine-generated text tend to overfit to their training distribution of data or source models.

Other works have trained supervised models for {\probfull} on top of neural representations \citep{bakhtin2019learning,release-strategies,uchendu-etal-2020-authorship,ippolito-etal-2020-automatic,tweepfake}, bag-of-words features \citep{release-strategies,tweepfake}, and handcrafted statistical features \citep{gehrmann-etal-2019-gltr}. Alternatively, \citet{release-strategies} notes the surprising efficacy of a simple zero-shot method for {\probfull}, which thresholds a candidate passage based on its average log probability under the generative model, serving as a strong baseline for zero-shot {\probfull} in our work. In our work, we similarly use the generating model to detect its own generations in a zero shot manner, but through a different approach based on estimating local curvature of the log probability around the sample rather than the raw log probability of the sample itself. See \citet{Jawahar2020AutomaticDO} for a complete survey on {\probfull}. Other work explores watermarks for generated text \citep{kirchenbauer2023watermark}, which modify a model's generations to make them easier to detect. Our work does not assume text is generated with the goal of easy detection; {\name} detects text generated from publicly available LLMs using standard LLM sampling strategies.

The widespread use of LLMs has led to much other contemporaneous work on detecting LLM output. \citet{sadasivan2023can} show that the detection AUROC of the an detector is upper bounded by a function of the TV distance between the model and human text. However, we find that AUROC of {\name} is high even for the largest publicly-available models (Table~\ref{tab:gpt-3-results}), suggesting that TV distance may not correlate strongly with model scale and capability. This disconnect may be exacerbated by new training objectives other than maximum likelihood, e.g., reinforcement learning with human feedback \citep{christiano2017deep,ziegler2020finetuning}. Both \citet{sadasivan2023can} and \citet{krishna2023paraphrasing} show the effectiveness of paraphrasing as a tool for evading detection, suggesting an important area of study for future work. \citet{liang2023gpt} show that multi-lingual detection is difficult, with non-{\name} detectors showing bias against non-native speakers; this result highlights the advantage of zero-shot detectors like {\name}, which generalize well to any data generated by the original generating model. \citet{mireshghallah2023smaller} study which proxy scoring models produce the most useful log probabilities for detection when the generating model is not known (a large-scale version of our Figure~\ref{fig:cross-eval}). Surprisingly (but consistent with our findings), they find that smaller models are in fact \textit{better} proxy models for performing detection with perturbation-based methods like {\name}.

The problem of {\probfull} echoes earlier work on detecting deepfakes, artificial images or videos generated by deep nets, which has spawned substantial efforts in detection of fake visual content \citep{DFDC2020,zi2020wilddeepfake}. While early works in deepfake detection used relatively general-purpose model architectures \citep{guera2018deepfake}, many deepfake detection methods rely on the continuous nature of image data to achieve state-of-the-art performance \citep{zhao2021multi,Guarnera_2020_CVPR_Workshops}, making direct application to text difficult.

\section{The Zero-Shot {\probfulltitle} Problem}
We study zero-shot {\probfull}, the problem of detecting whether a piece of text, or \textit{candidate passage} $x$, is a sample from a \textit{source model} $\sm$. The problem is zero-shot in the sense that we do not assume access to human-written or generated samples to perform detection. As in prior work, we study a `white box' setting \citep{gehrmann-etal-2019-gltr} in which the detector may evaluate the log probability of a sample $\log \sm(x)$. The white box setting \textbf{does not} assume access to the model architecture or parameters. Most public APIs for LLMs (such as GPT-3) enable scoring text, though some exceptions exist, notably ChatGPT. While most of our experiments consider the white box setting, see Section~\ref{sec:variants} for experiments in which we score text using models other than the source model. See \citet{mireshghallah2023smaller} for a comprehensive evaluation in this setting.

The detection criterion we propose, {\name}, also makes use of generic pre-trained mask-filling models in order to generate passages that are `nearby' the candidate passage. However, these mask-filling models are used off-the-shelf, without any fine-tuning or adaptation to the target domain.

\section{{\name}: Zero-shot {\probfulltitle} with Random Perturbations}
\label{sec:local-extremum}

\newlength{\textfloatsepsave}
\setlength{\textfloatsepsave}{\textfloatsep}
\setlength{\textfloatsep}{0pt}
\begin{algorithm}
    \caption{{\name} model-generated text detection}
    \label{alg:method}
    \begin{algorithmic}[1]
        \footnotesize \STATE \textbf{Input:} passage $x$, source model $\sm$, perturbation function $q$, number of perturbations $k$, decision threshold $\epsilon$
        \STATE $\tilde x_i \sim q(\cdot\mid x),\,i\in [1..k]$ \hfill\com{mask spans, sample replacements}
        \STATE $\tilde \mu \gets \frac{1}{k}\sum_i\log \sm(\tilde x_i)$ \hfill \com{approximate expectation in Eq. 1}
        \STATE $\pdh_x \gets \log \sm(x) - \tilde \mu$ \hfill\com{estimate $\pd$}
        \STATE \scalebox{0.95}{$\tilde \sigma_x^2 \gets \frac{1}{k-1}\sum_i\left(\log \sm(\tilde x_i) - \tilde \mu\right)^2$ \hfill\com{variance for normalization}}
        \IF{$\frac{\pdh_x}{\sqrt{\tilde \sigma_x}} > \epsilon$}
            \STATE return \texttt{true} \hfill\com{probably model sample}
        \ELSE
            \STATE return \texttt{false} \hfill\com{probably not model sample}
        \ENDIF
    \end{algorithmic}
\end{algorithm}
\setlength{\textfloatsep}{\textfloatsepsave}

{\name} is based on the hypothesis that samples from a source model $\sm$ typically lie in areas of negative curvature of the log probability function of $\sm$, unlike human text. In other words, if we apply small perturbations to a passage $x\sim\sm$, producing $\tilde x$, the quantity $\log \sm(x) - \log \sm(\tilde x)$ should be relatively large on average for machine-generated samples compared to human-written text. To leverage this hypothesis, first consider a perturbation function $q(\cdot\mid x)$ that gives a distribution over $\tilde x$, slightly modified versions of $x$ with similar meaning (we will generally consider roughly paragraph-length texts $x$). As an example, $q(\cdot\mid x)$ might be the result of simply asking a human to rewrite one of the sentences of $x$, while preserving the meaning of $x$. Using the notion of a perturbation function, we can define the \textit{perturbation discrepancy} $\pd$: 
\begin{equation}
    \pd \triangleq \log \sm(x) - \mathbb{E}_{\tilde x \sim q(\cdot\mid x)} \log \sm(\tilde x)
    \label{eq:perturbation-discrepancy}
\end{equation}
We state our hypothesis more formally as the Local Perturbation Discrepancy Gap Hypothesis, which describes a gap in the perturbation discrepancy for model-generated text and human-generated text.

\noindent\textbf{Perturbation Discrepancy Gap Hypothesis.}
\textit{If $q$ produces samples on the data manifold, $\pd$ is positive and large with high probability for samples $x\sim\sm$. For human-written text, $\pd$ tends toward zero for all $x$.}

\begin{figure}
    \centering 
    \includegraphics[width=\columnwidth]{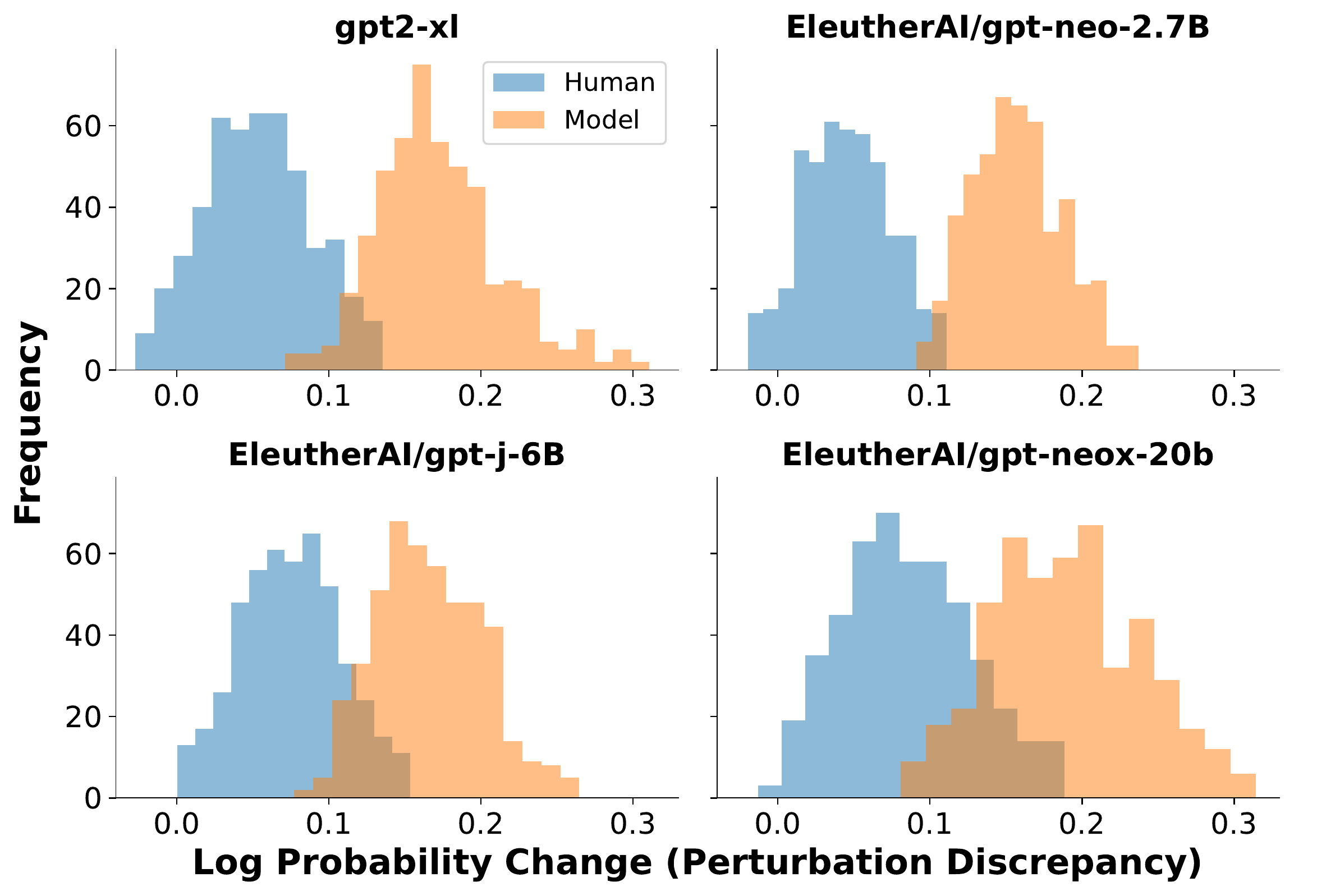}
    \vspace{-5mm}
    \caption{The average drop in log probability (perturbation discrepancy) after rephrasing a passage is consistently higher for model-generated passages than for human-written passages. Each plot shows the distribution of the perturbation discrepancy $\pd$ for \textcolor{human-blue}{\textbf{human-written news articles}} and \textcolor{machine-orange}{\textbf{machine-generated articles}} of equal word length. Human-written articles are a sample of 500 XSum articles; machine-generated text, generated from models GPT-2 (1.5B), GPT-Neo-2.7B \citep{gpt-neo}, GPT-J (6B; \citet{gpt-j}) and GPT-NeoX (20B; \citet{gpt-neox-20b}), is generated by prompting each model with the first 30 tokens of each XSum article, sampling from the raw conditional distribution. Discrepancies are estimated with 100 T5-3B samples.}
    \vspace{-2mm}
    \label{fig:perturbation-discrepancy}
\end{figure}

If we define $q(\cdot\mid x)$ to be samples from a mask-filling model such as T5 \citep{raffel2020t5}, rather than human rewrites, we can empirically test the Perturbation Discrepancy Gap Hypothesis in an automated, scalable manner. For real data, we use 500 news articles from the XSum dataset \citep{shashi2018dont}; for model samples, we use the output of four different LLMs when prompted with the first 30 tokens of each article in XSum. We use T5-3B to apply perturbations, masking out randomly-sampled 2-word spans until 15\% of the words in the article are masked. We approximate the expectation in Eq.~\ref{eq:perturbation-discrepancy} with 100 samples from T5.\footnote{We later show in Figure~\ref{fig:n-perturb} that varying the number of samples used to estimate the expectation effectively allows for trading off between accuracy and speed.} Figure~\ref{fig:perturbation-discrepancy} shows the result of this experiment. We find the distribution of perturbation discrepancies is significantly different for human-written articles and model samples; model samples tend to have a larger perturbation discrepancy. Section~\ref{sec:ablations} explores a relaxation of the assumption that $q$ only produces samples on the data manifold, finding that a gap, although reduced, still exists in this case.

Given these results, we can detect if a piece of text was generated by a model $\sm$ by simply thresholding the perturbation discrepancy. In practice, we find that normalizing the perturbation discrepancy by the standard deviation of the observed values used to estimate $\mathbb{E}_{\tilde x \sim q(\cdot\mid x)} \log \sm(\tilde x)$ provides a slightly better signal for detection, typically increasing AUROC by around 0.020, so we use this normalized version of the perturbation discrepancy in our experiments. The resulting method, {\name}, is summarized in Alg.~\ref{alg:method}. Having described an application of the perturbation discrepancy to {\probfull}, we next provide an interpretation of this quantity.



\textbf{Interpretation of perturbation discrepancy as curvature}


While Figure~\ref{fig:perturbation-discrepancy} suggests that the perturbation discrepancy may be useful, it is not immediately obvious what it measures. In this section, we show that
the perturbation discrepancy approximates a measure of the local curvature of the log probability function near the candidate passage, more specifically, that it is proportional to the negative trace of the Hessian of the log probability function.\footnote{Rather than the Hessian of the log likelihood with respect to model parameters (the Fisher Information Matrix), here we refer to the Hessian of the log probability with respect to the sample $x$.} To handle the non-differentiability of discrete data, we consider candidate passages in a latent semantic space, where small displacements correspond to valid edits that retain similar meaning to the original. Because our perturbation function (T5) models natural text, we expect our perturbations to roughly capture such meaningful variations of the original passage, rather than arbitrary edits.

We first invoke Hutchinson's trace estimator \citep{hutchinson1990stochastic}, giving an unbiased estimate of the trace of matrix $A$:
\begin{equation}
    \text{tr}(A) = \mathbb{E}_\mathbf{z} \mathbf{z}^\top A \mathbf{z}
    \label{eq:hutch}
\end{equation}
provided that the elements of $\mathbf{z} \sim q_z$ are IID with $\mathbb{E}[z_i] = 0$ and $\text{Var}(z_i) = 1$. To use Equation~\ref{eq:hutch} to estimate the trace of the Hessian of $f$ at $x$, we must therefore compute the expectation of the directional second derivative $\mathbf{z}^\top H_f(x)\mathbf{z}$. We approximate this expression with finite differences:
\begin{equation}
    \mathbf{z}^\top H_f(x) \mathbf{z} \approx \frac{f(x + h\mathbf{z}) + f(x - h\mathbf{z}) - 2f(x)}{h^2}
    \label{eq:fd}
\end{equation}
Combining Equations~\ref{eq:hutch} and~\ref{eq:fd} and simplifying with $h=1$, we have an estimate of the negative Hessian trace
\begin{align}
    -\text{tr}\left(H_f(x)\right) &\approx 2f(x) - \mathbb{E}_\mathbf{z} \left[f(x + \mathbf{z}) + f(x - \mathbf{z})\right].
    \label{eq:neg-trace}
\end{align}
If our noise distribution is \textit{symmetric}, that is, $p(\mathbf{z}) = p(-\mathbf{z})$ for all $\mathbf{z}$, then we can simplify Equation~\ref{eq:neg-trace} to
\begin{align}
    \frac{-\text{tr}\left(H_f(x)\right)}{2} &\approx f(x) - \mathbb{E}_\mathbf{z} f(x + \mathbf{z}).
    \label{eq:simplified}
\end{align}

\addtolength{\tabcolsep}{-0.3em}
\begin{table*}
    \centering
    \small
    \resizebox{\textwidth}{!}{%
    \begin{tabular}{lcccccc|cccccc|cccccc}
         & \multicolumn{6}{c}{\textbf{XSum}} & \multicolumn{6}{c}{\textbf{SQuAD}} & \multicolumn{6}{c}{\textbf{WritingPrompts}} \\
         \cmidrule(lr){2-7} \cmidrule(lr){8-13} \cmidrule(lr){14-19}
         \textbf{Method} & GPT-2 & OPT-2.7 & Neo-2.7 & GPT-J & NeoX & \textbf{Avg.} & GPT-2 & OPT-2.7 & Neo-2.7 & GPT-J & NeoX & \textbf{Avg.} & GPT-2 & OPT-2.7 & Neo-2.7 & GPT-J & NeoX & \textbf{Avg.} \\
         \midrule
         $\log p(x)$ & 0.86 & 0.86 & 0.86 & 0.82 & 0.77 & 0.83 & 0.91 & 0.88 & 0.84 & 0.78 & 0.71 & 0.82 & 0.97 & 0.95 & 0.95 & 0.94 & \phantom{*}0.93* & 0.95 \\
        Rank & 0.79 & 0.76 & 0.77 & 0.75 & 0.73 & 0.76 & 0.83 & 0.82 & 0.80 & 0.79 & 0.74 & 0.80 & 0.87 & 0.83 & 0.82 & 0.83 & 0.81 & 0.83 \\
        LogRank & \phantom{*}0.89* & \phantom{*}0.88* & \phantom{*}0.90* & \phantom{*}0.86* & \phantom{*}0.81* & \phantom{*}0.87* & \phantom{*}0.94* & \phantom{*}0.92* & \phantom{*}0.90* & \phantom{*}0.83* & \phantom{*}0.76* & \phantom{*}0.87* & \phantom{*}0.98* & \phantom{*}0.96* & \phantom{*}0.97* & \phantom{*}0.96* & \textbf{0.95} & \phantom{*}0.96* \\
        Entropy & 0.60 & 0.50 & 0.58 & 0.58 & 0.61 & 0.57 & 0.58 & 0.53 & 0.58 & 0.58 & 0.59 & 0.57 & 0.37 & 0.42 & 0.34 & 0.36 & 0.39 & 0.38 \\
        {\name} & \textbf{0.99} & \textbf{0.97} & \textbf{0.99} & \textbf{0.97} & \textbf{0.95} & \textbf{0.97} & \textbf{0.99} & \textbf{0.97} & \textbf{0.97} & \textbf{0.90} & \textbf{0.79} & \textbf{0.92} & \textbf{0.99} & \textbf{0.99} & \textbf{0.99} & \textbf{0.97} & \phantom{*}0.93* & \textbf{0.97} \\
        \midrule
        Diff & 0.10 & 0.09 & 0.09 & 0.11 & 0.14 & 0.10 & 0.05 & 0.05 & 0.07 & 0.07 & 0.03 & 0.05 & 0.01 & 0.03 & 0.02 & 0.01 & -0.02 & 0.01 \\
         \bottomrule
    \end{tabular}}
    \vspace{-3mm}
    \caption{AUROC for detecting samples from the given model on the given dataset for {\name} and four previously proposed criteria (500 samples used for evaluation). From 1.5B parameter GPT-2 to 20B parameter GPT-NeoX, {\name} consistently provides the most accurate detections. \textbf{Bold} shows the best AUROC within each column (model-dataset combination); asterisk (*) denotes the second-best AUROC. Values in the final row show {\name}'s AUROC over the strongest baseline method in that column.}
    \label{tab:main-results}
    \vspace{-3mm}
\end{table*}
\addtolength{\tabcolsep}{0.3em}

We note that the RHS of Equation~\ref{eq:simplified} corresponds to the perturbation discrepancy (\ref{eq:perturbation-discrepancy}) where the perturbation function $q(\tilde{x} \mid x)$ is replaced by the distribution $q_z(z)$ used in Hutchinson's trace estimator (\ref{eq:hutch}).
Here, $\tilde{x}$ is a high-dimensional sequence of tokens while $q_z$ is a vector in a compact semantic space.
Since the mask-filling model samples sentences similar to $x$ with minimal changes to semantic meaning, we can think of the mask-filling model as first sampling a similar semantic embedding ($\tilde{z} \sim q_z$) and then mapping this to a token sequence ($\tilde{z} \mapsto \tilde{x}$).
Sampling in semantic space ensures that all samples stay near the data manifold, which is useful because we would expect the log probability to always drop if we randomly perturb tokens.
We can therefore interpret our objective as approximating the curvature restricted to the data manifold.

\section{Experiments}

We conduct experiments to better understand multiple facets of {\probfull}; we study the effectiveness of {\name} for zero-shot {\probfull} compared to prior zero-shot approaches, the impact of distribution shift on zero-shot and supervised detectors, and detection accuracy for the largest publicly-available models. To further characterize factors that impact detection accuracy, we also study the robustness of zero-shot methods to machine-generated text that has been partially revised, the impact of alternative decoding strategies on detection accuracy, and a black-box variant of the detection task. Finally, we analyze more closely {\name}'s behavior as the choice of perturbation function, the number of samples used to estimate $\pd$, the length of the passage, and the data distribution is varied.

\textbf{Comparisons.} We compare {\name} with various existing zero-shot methods for {\probfull} that also leverage the predicted token-wise conditional distributions of the source model for detection. These methods correspond to statistical tests based on token log probabilities, token ranks, or predictive entropy \citep{gehrmann-etal-2019-gltr,release-strategies,ippolito-etal-2020-automatic}. The first method uses the source model's average token-wise log probability to determine if a candidate passage is machine-generated or not; passages with high average log probability are likely to be generated by the model. The second and third methods use the average observed rank or log-rank of the tokens in the candidate passage according to the model's conditional distributions. Passages with smaller average (log-)rank are likely machine-generated. We also evaluate an entropy-based approach inspired by the hypothesis in \citet{gehrmann-etal-2019-gltr} that model-generated texts will be more `in-distribution' for the model, leading to more over-confident (thus lower entropy) predictive distributions. Empirically, we find predictive entropy to be \textit{positively} correlated with passage fake-ness more often that not; therefore, this baseline uses high average entropy in the model's predictive distribution as a signal that a passage is machine-generated. While our main focus is on zero-shot detectors as they do not require re-training for new domains or source models, for completeness we perform comparisons to supervised detection models in Section~\ref{sec:main-results}, using OpenAI's RoBERTa-based \citep{liu2019roberta} GPT-2 detector models,\footnote{\href{https://github.com/openai/gpt-2-output-dataset/tree/master/detector}{\texttt{https://github.com/openai/gpt-2-output-\\dataset/tree/master/detector}}} which are fine-tuned on millions of samples from various GPT-2 model sizes and decoding strategies.

\textbf{Datasets \& metrics} Our experiments use six datasets that cover a variety of everyday domains and LLM use-cases. We use news articles from the XSum dataset \citep{shashi2018dont} to represent fake news detection, Wikipedia paragraphs from SQuAD contexts \citep{rajpurkar-etal-2016-squad} to represent machine-written academic essays, and prompted stories from the Reddit WritingPrompts dataset \citep{fan-etal-2018-hierarchical} to represent detecting machine-generated creative writing submissions. To evaluate robustness to distribution shift, we also use the English and German splits of WMT16 \citep{bojar2016wmt} as well as long-form answers written by human experts in the PubMedQA dataset \citep{jin-etal-2019-pubmedqa}. Each experiment uses between 150 and 500 examples for evaluation, as noted in the text. For each experiment, we generate the machine-generated text by prompting with the first 30 tokens of the real text (or just the question tokens for the PubMedQA experiments). We measure performance using the area under the receiver operating characteristic curve (AUROC), which can be interpreted as the probability that a classifier correctly ranks a randomly-selected positive (machine-generated) example higher than a randomly-selected negative (human-written) example. All experiments use an equal number of positive and negative examples.

\textbf{Hyperparameters.} The key hyperparameters of {\name} are the fraction of words masked for perturbation, the length of the masked spans, the model used for mask filling, and the sampling hyperparameters for the mask-filling model. Using BERT \citep{devlin-etal-2019-bert} masked language modeling as inspiration, we use 15\% as the mask rate. We performed a small sweep over masked span lengths of $\{2,5,10\}$ on a held-out set of XSum data, finding 2 to perform best. We use these settings for \textbf{all experiments, without re-tuning}. We use T5-3B for almost all experiments, except for GPT-NeoX and GPT-3 experiments, where compute resources allowed for the larger T5-11B model; we also use mT5-3B instead of T5-3B for the WMT multilingual experiment. We do not tune the hyperparameters for the mask filling model, sampling directly with temperature 1.

\begin{figure}
    \centering
    \includegraphics[width=\columnwidth]{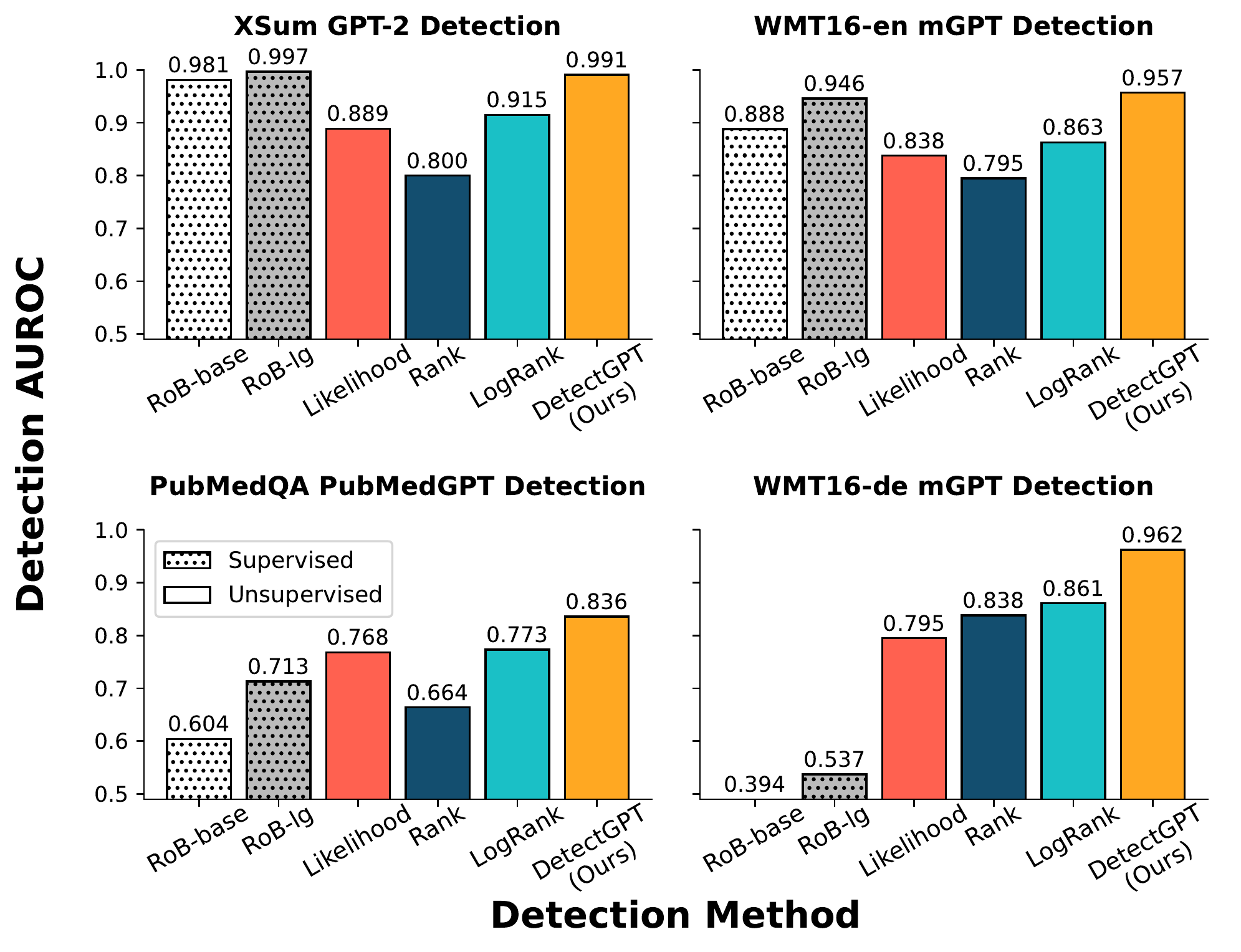}
    \vspace{-5mm}
    \caption{Supervised machine-generated text detection models trained on large datasets of real and generated texts perform as well as or better than {\name} on \textbf{in-distribution (top row)} text. However, zero-shot methods work out-of-the-box for \textbf{new domains (bottom row)} such as PubMed medical texts and German news data from WMT16. For these domains, supervised detectors fail due to excessive distribution shift.}
    \vspace{-5mm}
    \label{fig:supervised}
\end{figure}

\subsection{Main Results}
\label{sec:main-results}

We first present two groups of experiments to evaluate {\name} along with existing methods for zero-shot and supervised detection on models from 1.5B to 175B parameters.

\textbf{Zero-shot machine-generated text detection.} We present the comparison of different zero-shot detection methods in Table \ref{tab:main-results}. In these experiments, model samples are generated by sampling from the raw conditional distribution with temperature 1. {\name} most improves average detection accuracy for XSum stories (0.1 AUROC improvement) and SQuAD Wikipedia contexts (0.05 AUROC improvement). While it also performs accurate detection for WritingPrompts, the performance of all methods tends to increase, and the average margin of improvement is narrow.\footnote{The overall ease of detecting machine-generated fake writing corroborates anecdotal reporting that machine-generated creative writing tends to be noticeably generic, and therefore relatively easy to detect \citep{roose2022chatgpt}.} For 14 of the 15 combinations of dataset and model, {\name} provides the most accurate detection performance, with a 0.06 AUROC improvement on average. Log-rank thresholding proves to be a consistently stronger baseline than log probability thresholding, although it requires slightly more information (full predicted logits), which are not always available in public APIs.

\textbf{Comparison with supervised detectors.}
While our experiments generally focus on zero-shot detection, some works have evaluated the detection performance of supervised methods (typically fine-tuned transformers) for detecting machine-generated text. In this section, we explore several domains to better understand the relative strengths of supervised and zero-shot detectors. The results are presented in Figure~\ref{fig:supervised}, using 200 samples from each dataset for evaluation. We find that supervised detectors can provide similar detection performance to {\name} on \textit{in-distribution} data like English news, but perform significantly worse than zero-shot methods in the case of English scientific writing and fail altogether for German writing. This finding echoes past work showing that language models trained for machine-generated text detection overfit to their training data (source model, decoding strategy, topic, language, etc.; \citet{uchendu-etal-2020-authorship,ippolito-etal-2020-automatic,Jawahar2020AutomaticDO}). In contrast, zero-shot methods generalize relatively easily to new languages and domains; {\name}'s performance in particular is mostly unaffected by the change in language from English to German.


\begin{table}
    \small
    \centering
    \begin{tabular}{lccc|c}
        \toprule
         & \textbf{PMQA} & \textbf{XSum} & \textbf{WritingP} & \textbf{Avg.}  \\
         \midrule
         RoB-base & 0.64 / 0.58 & \textbf{0.92} / 0.74 & \textbf{0.92} / 0.81 & 0.77 \\
         RoB-large & 0.71 / 0.64 & \textbf{0.92} / \textbf{0.88} & 0.91 / \textbf{0.88} & \textbf{0.82} \\
         \midrule
         $\log p(x)$ & 0.64 / 0.55 & 0.76 / 0.61 & 0.88 / 0.67 & 0.69 \\
         \name & \textbf{0.84} / \textbf{0.77} & 0.84 / 0.84 & 0.87 / 0.84 & \textbf{0.83} \\
         \bottomrule
    \end{tabular}
    \caption{{\name} detects generations from GPT-3 and Jurassic-2 Jumbo (175B models from OpenAI and AI21 Labs) with average AUROC on-par with supervised models trained specifically for machine-generated text detection. For more `typical' text, such as news articles, supervised methods perform strongly. The GPT-3 AUROC appears first in each column, the Jurassic-2 AUROC appears second (i.e., after the slash).}
    \vspace{-5mm}
    \label{tab:gpt-3-results}
\end{table}

While our experiments have shown that {\name} is effective on a variety of domains and models, it is natural to wonder if it is effective for the largest publicly-available LMs. Therefore, we also evaluate multiple zero-shot and supervised methods on two 175B parameter models, OpenAI's GPT-3 and AI21 Labs' Jurassic-2 Jumbo. Because neither API provides access to the complete conditional distribution for each token, we cannot compare to the rank, log rank, and entropy-based prior methods. We sample 150 examples\footnote{We reduce the number of evaluation samples from 500 in our main experiments to reduce the API costs of these experiments.}
from the PubMedQA, XSum, and WritingPrompts datasets and compare the two pre-trained RoBERTa-based detector models with {\name} and the probability thresholding baseline. We show in Table~\ref{tab:gpt-3-results} that {\name} can provide detection competitive with or better than the stronger of the two supervised models, and it again greatly outperforms probability thresholding on average.

\begin{figure}
    \centering
    \includegraphics[width=0.9\columnwidth]{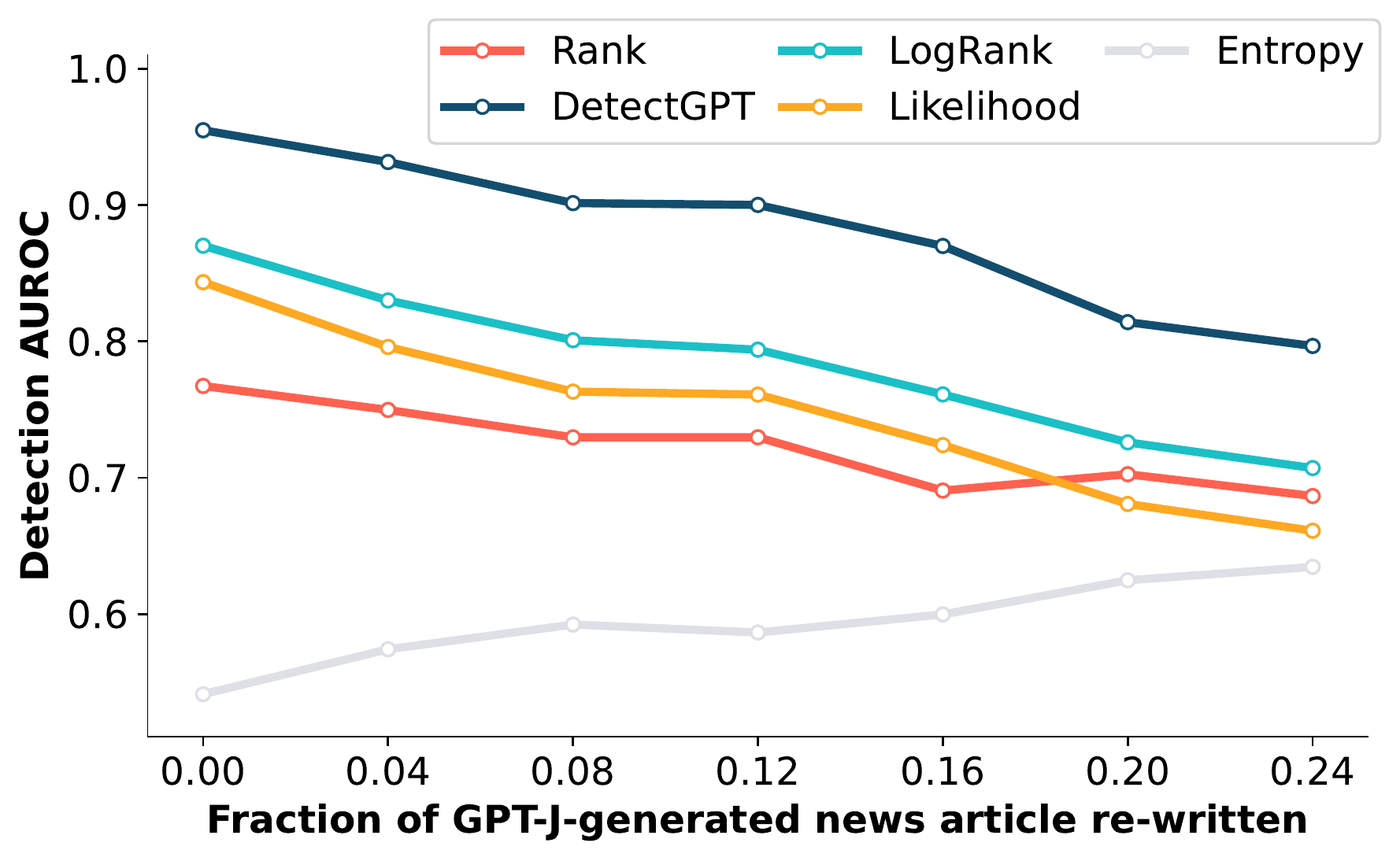}
    \vspace{-2mm}
    \caption{We simulate human edits to machine-generated text by replacing varying fractions of model samples with T5-3B generated text (masking out random five word spans until $r$\% of text is masked to simulate human edits to machine-generated text). The four top-performing methods all generally degrade in performance with heavier revision, but {\name} is consistently most accurate. Experiment is conducted on the XSum dataset.}
    \label{fig:pre-perturb}
\end{figure}

\subsection{Variants of {\probfulltitle}}
\label{sec:variants}
\textbf{Detecting paraphrased machine-generated text.} In practice, humans may manually edit or refine machine-generated text rather than blindly use a model's generations for their task of interest. We therefore conduct an experiment to simulate the detection problem for model samples that have been increasingly heavily revised. We simulate human revision by replacing 5 word spans of the text with samples from T5-3B until $r$\% of the text has been replaced, and report performance as $r$ varies. Figure~\ref{fig:pre-perturb} shows that {\name} maintains detection AUROC above 0.8 even when nearly a quarter of the text in model samples has been replaced. Unsurprisingly, almost all methods show a gradual degradation in performance as the sample is more heavily revised. The entropy baseline shows surprisingly robust performance in this setting (althought it is least accurate on average), even slightly improving detection performance up to 24\% replacement.
{\name} shows the strongest detection performance for all revision levels.

\textbf{Impact of alternative decoding strategies on detection.} While Table~\ref{tab:main-results} suggests that {\name} is effective for detecting machine-generated text, prior work notes that the decoding strategy (i.e., temperature sampling, top-$k$, nucleus/top-$p$) can impact the difficulty of detection. We repeat the analysis from Section~\ref{sec:main-results} using top-$k$ sampling and nucleus sampling. Top-$k$ sampling truncates the sampling distribution to only the $k$ highest-probability next tokens; nucleus sampling samples from only the smallest set of tokens whose combined probability exceeds $p$. The results are summarized in Table~\ref{tab:decoding}; Appendix Tables~\ref{tab:main-results-topp} and~\ref{tab:main-results-topk} show complete results. We use $k=40$, and $p=0.96$, in line with prior work \citep{ippolito-etal-2020-automatic}. We find that both top-$k$ and nucleus sampling make detection easier, on average. Averaging across domains, {\name} provides the clearest signal for zero-shot detection.

\addtolength{\tabcolsep}{-0.2em}
\begin{table}
    \footnotesize
    \centering
    \begin{tabular}{lcccccc}
    \toprule
        & \multicolumn{2}{c}{XSum} & \multicolumn{2}{c}{SQuAD} & \multicolumn{2}{c}{WritingPrompts} \\
        \cmidrule(lr){2-3} \cmidrule(lr){4-5} \cmidrule(lr){6-7}
        Method & top-$p$ & top-$k$ & top-$p$ & top-$k$ & top-$p$ & top-$k$ \\
        \midrule
        $\log p(x)$ & 0.92 & 0.87 & 0.89 & 0.85 & \textbf{0.98} & 0.96 \\
        Rank & 0.76 & 0.76 & 0.81 & 0.80 & 0.84 & 0.83 \\
        LogRank & \phantom{*}0.93* & \phantom{*}0.90* & \phantom{*}0.92* & \phantom{*}0.90* & \textbf{0.98} & \textbf{0.97} \\
        Entropy & 0.53 & 0.55 & 0.54 & 0.56 & 0.32 & 0.35 \\
        \midrule
        {\name} & \textbf{0.98} & \textbf{0.98} & \textbf{0.94} & \textbf{0.93} & \textbf{0.98} & \textbf{0.97} \\
        \bottomrule
    \end{tabular}
    \caption{AUROC for zero-shot methods averaged across the five models in Table~\ref{tab:main-results} for both top-$k$ and top-$p$ sampling, with $k=40$ and $p=0.96$. Both settings enable slightly more accurate detection, and {\name} consistently provides the best detection performance. See Appendix Tables~\ref{tab:main-results-topp} and~\ref{tab:main-results-topk} for complete results.}
    \vspace{-4mm}
    \label{tab:decoding}
\end{table}
\addtolength{\tabcolsep}{0.2em}

\textbf{Detection when the source model is unknown.} While our experiments have focused on the white-box setting for {\probfull}, in this section, we \begin{wrapfigure}{r}{3.8cm}
    \vspace{-1mm}
    \centering
    \includegraphics[width=3.8cm]{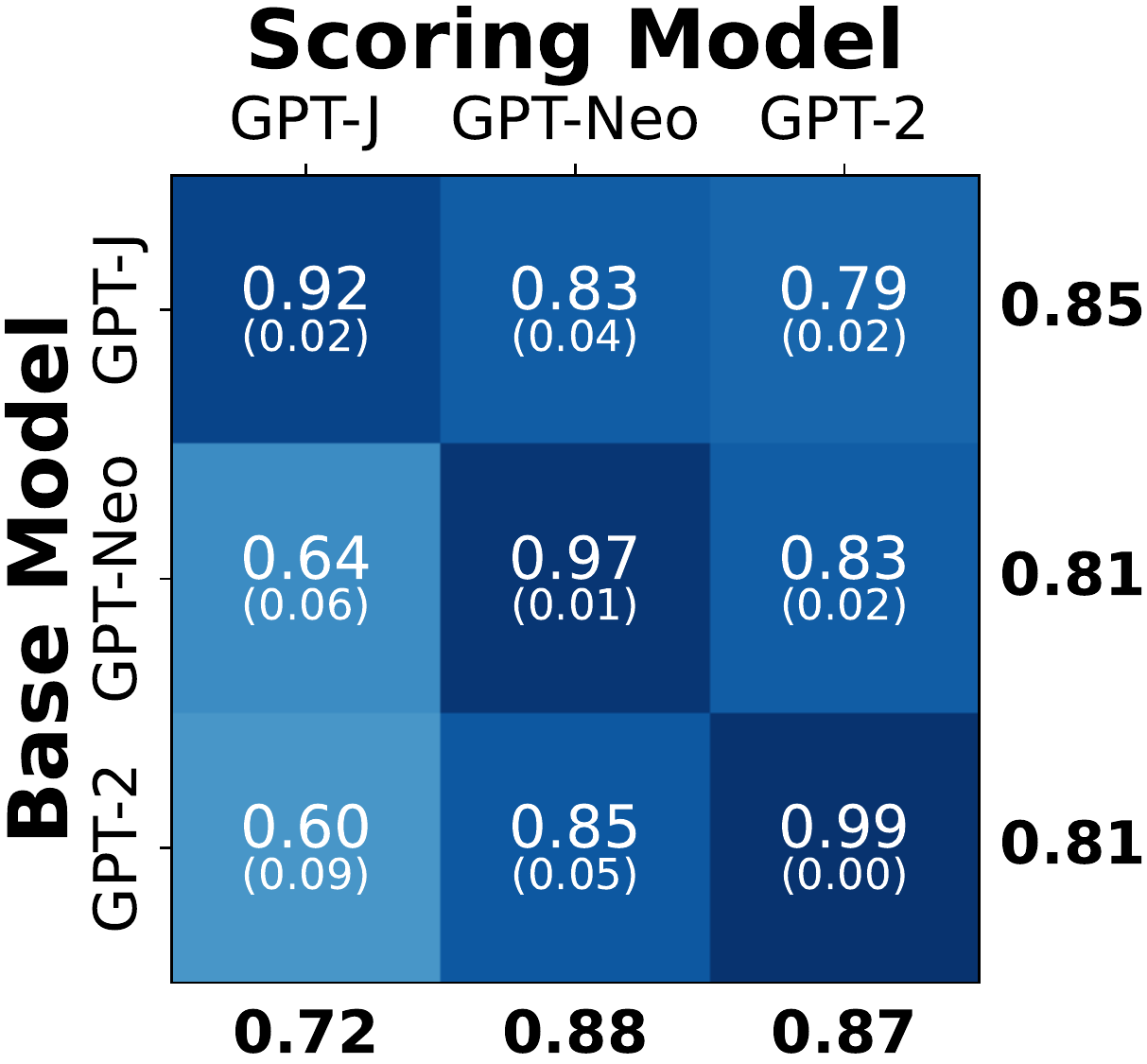}
    \vspace{-5mm}
    \caption{{\name} performs best when scoring samples with the same model that generated them (diagonal), but the column means suggest that some models (GPT-Neo, GPT-2) may be better `scorers' than others (GPT-J). White values show mean (standard error) AUROC over XSum, SQuAD, and WritingPrompts; \textbf{black} shows row/column mean.}
    \vspace{-2mm}
    \label{fig:cross-eval}
\end{wrapfigure} explore the effect of using a different model to \textit{score} a candidate passage (and perturbed texts) than the model that generated the passage. In other words, we aim to classify between human-generated text and text from model $A$, but without access to model $A$ to compute log probabilities. Instead, we use log probabilities computed by a surrogate model $B$. We consider three models, GPT-J, GPT-Neo-2.7, and GPT-2, evaluating all possible combinations of source model and surrogate model (9 total). We average the performance across 200 samples from XSum, SQuAD, and WritingPrompts. The results are presented in Figure~\ref{fig:cross-eval}, showing that when the surrogate model is different from the source model, detection performance is reduced, indicating that {\name} is most suited to the white-box setting. Yet we also observe that if we fix the model used for scoring and average across source models whose generations are detected (average within column), there is significant variation in AUROC; GPT-2 and GPT-Neo-2.7 seem to be better `scorers' than GPT-J. These variations in cross-model scoring performance suggest ensembling scoring models may be a useful direction for future research; see \citet{mireshghallah2023smaller} for reference.

\subsection{Other factors impacting performance of {\name}}
In this section, we explore how factors such as the size of the mask-filling model, the number of perturbations used to estimate the expectation in Equation~\ref{eq:perturbation-discrepancy}, or the data distribution of the text to be detected impact detection quality.

\label{sec:ablations}

\begin{figure}
    \centering
    \includegraphics[width=\columnwidth]{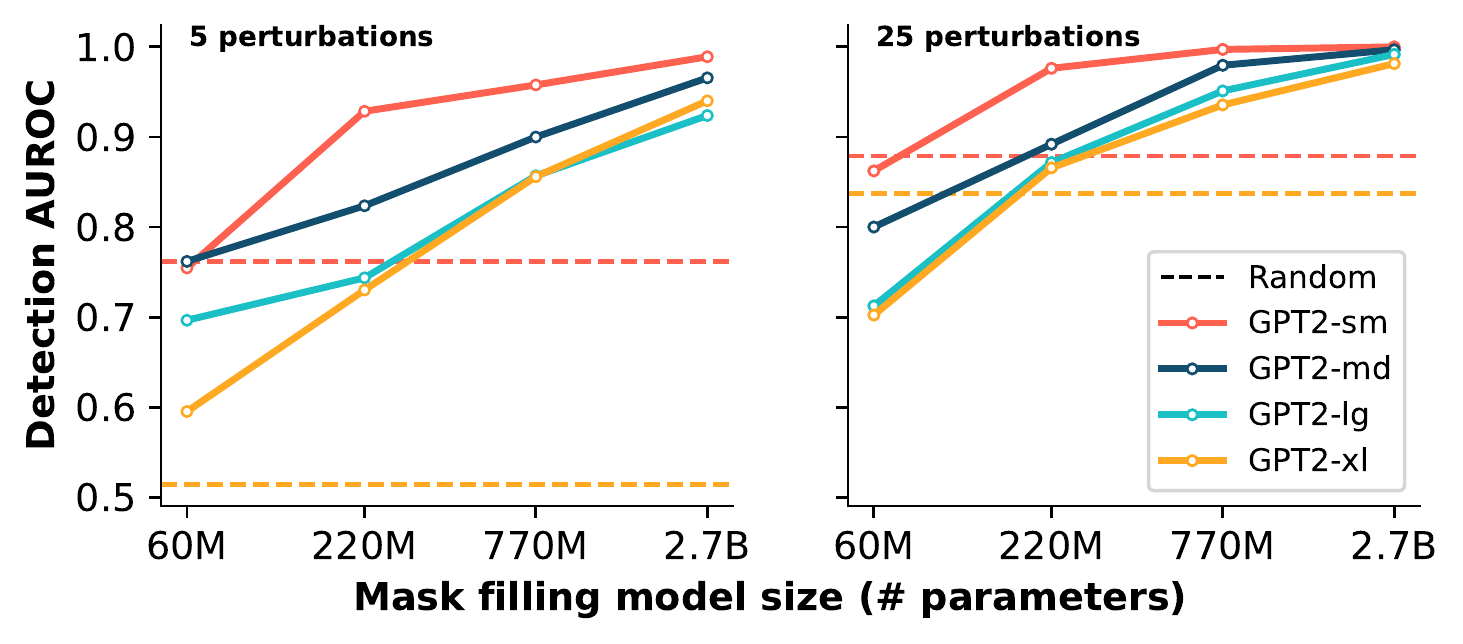}
    \vspace{-6mm}
    \caption{There is a clear association between capacity of mask-filling model and detection performance, across source model scales. Random mask filling (uniform sampling from mask filling model vocabulary) performs poorly, reinforcing the idea that the perturbation function should produce samples on the data manifold. Curves show AUROC scores on 200 SQuAD contexts.}
    \vspace{-5mm}
    \label{fig:model-size}
\end{figure}

\textbf{Source and mask-filling model scale.} Here we study the impact of the size of the source model and mask-filling model on {\name}'s performance; the results are shown in Figure~\ref{fig:model-size}. In particular, the increased discrimination power of {\name} for larger mask-filling models supports the interpretation that {\name} is estimating the curvature of the log probability in a latent semantic space, rather than in raw token embedding space. Larger T5 models better represent this latent space, where random directions correspond to meaningful changes in the text.

\begin{figure}
    \centering
    \includegraphics[width=\columnwidth]{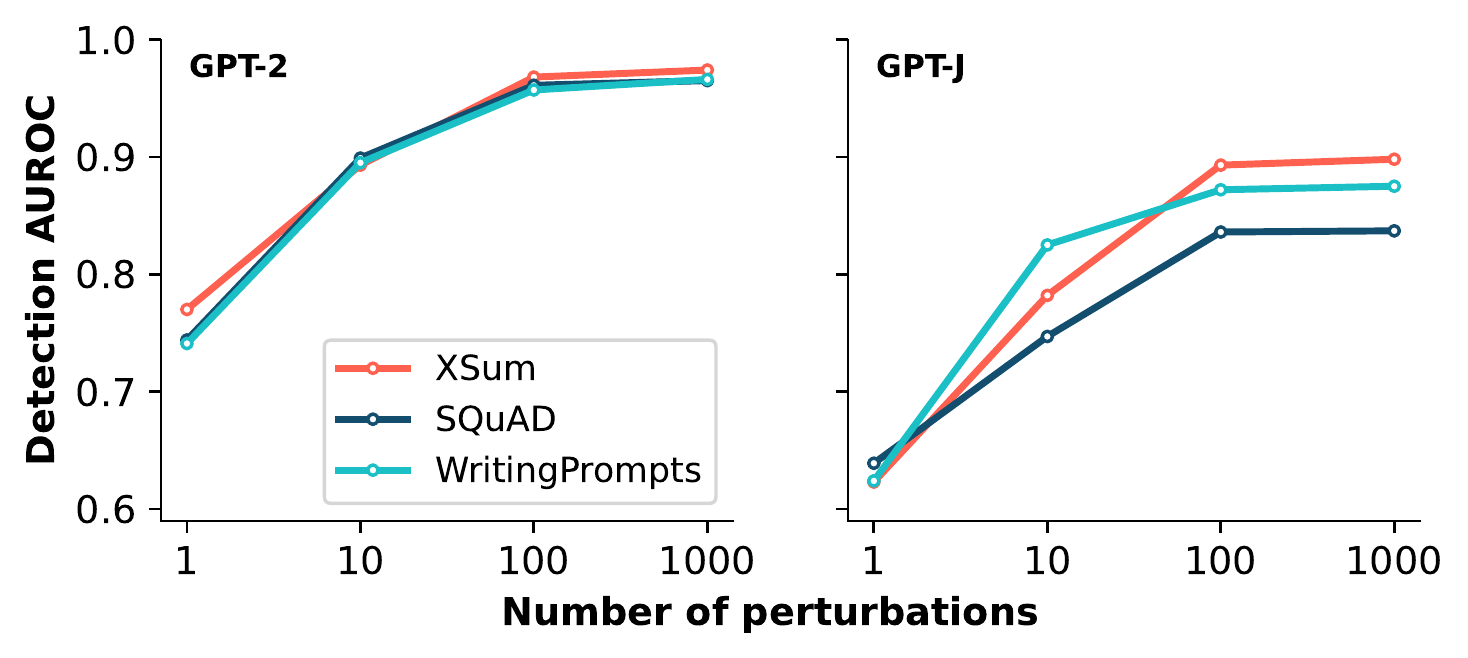}
    \vspace{-6mm}
    \caption{Impact of varying the number of perturbations (samples of mask and mask-fill) used by {\name} on AUROC for GPT-2 (\textbf{left}) and GPT-J (\textbf{right}) to estimate the perturbation discrepancy on detection. Averaging up to 100 perturbations greatly increases {\name}'s reliability. Perturbations sampled from T5-large.}
    \vspace{-4mm}
    \label{fig:n-perturb}
\end{figure}

\textbf{Number of perturbations for {\name}.} We evaluate the performance of {\name} as a function of the number of perturbations used to estimate the expectation in Equation~\ref{eq:perturbation-discrepancy} on three datasets. The results are presented in Figure~\ref{fig:n-perturb}. Detection accuracy continues to improve until 100 perturbations, where it converges. Evaluations use 100 examples from each dataset.

\begin{figure}
    \centering 
    \includegraphics[width=0.98\columnwidth]{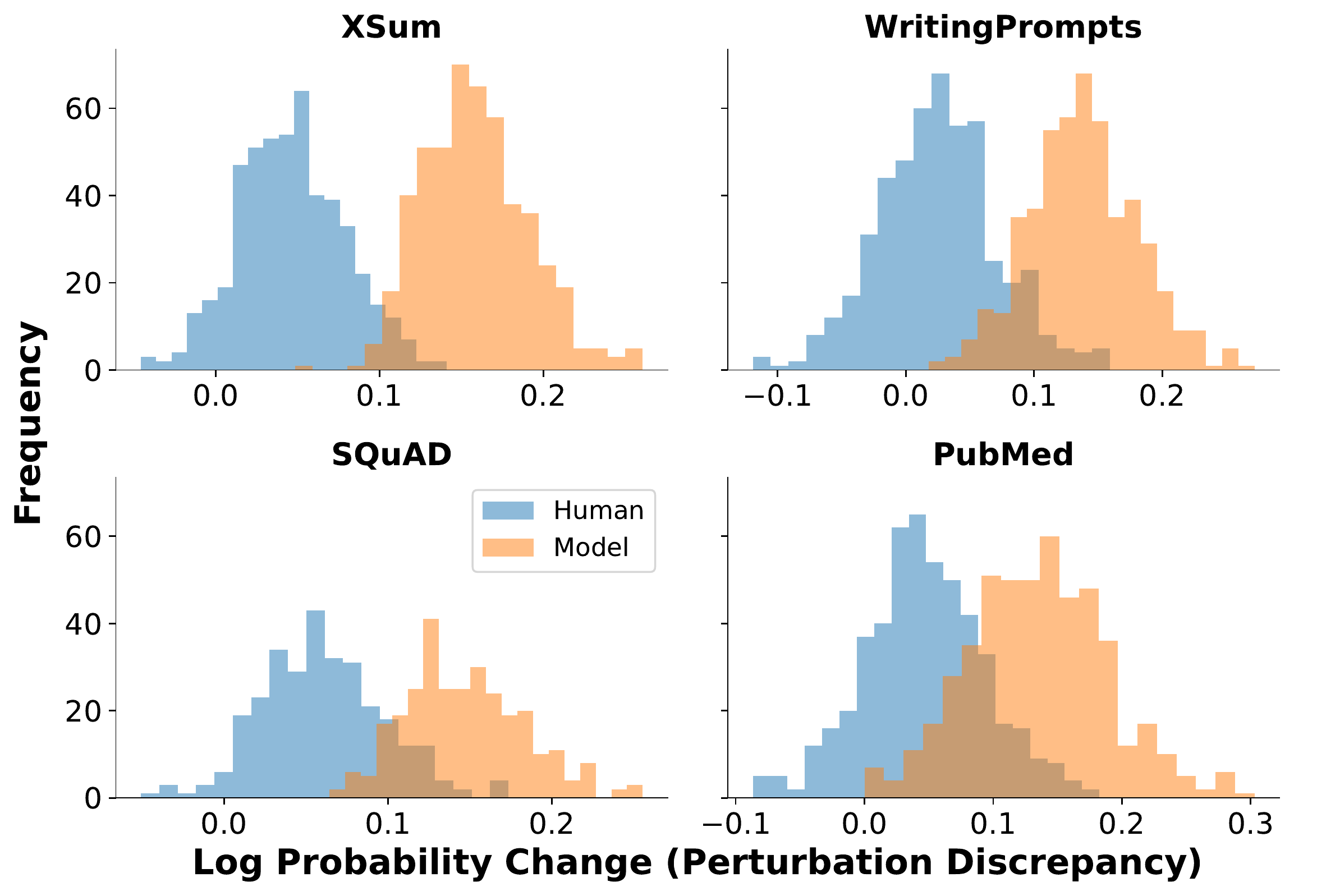}
    \caption{Perturbation discrepancy distributions for GPT-Neo (2.7B) and humans across domains. A threshold of 0.1 generally separates model- and human-generated text well, which is important for practical scenarios where the domain is unknown.}
    \vspace{-5mm}
    \label{fig:perturbation-discrepancy-neo}
\end{figure}

\begin{figure}
    \centering 
    \includegraphics[width=0.98\columnwidth]{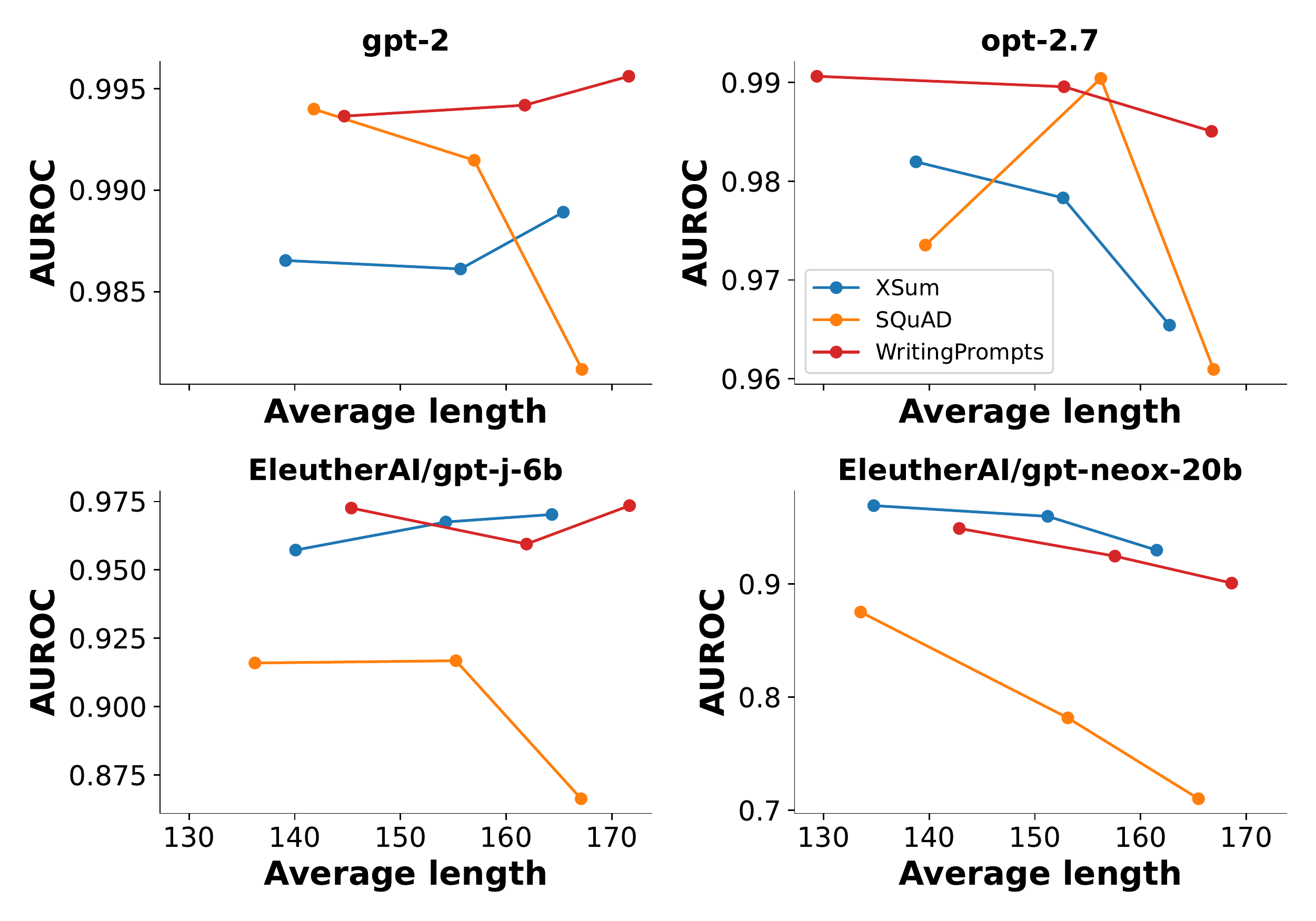}
    \vspace{-1.6mm}
    \caption{{\name} AUROC vs passage length. The relationship between detection performance and passage length generally depends on the dataset and model (or tokenizer). Decreases in detection quality with increasing length may be due to T5 failing to track many (20+) masks to fill at once; this problem may be mitigated by applying mask-fills in a sequence of smaller batches.}
    \vspace{-5.5mm}
    \label{fig:auroc-vs-len}
\end{figure}

\textbf{Data distributional properties.} We study more closely the impact of the data distribution on {\name}, particularly how the domain impacts the threshold separating the perturbation discrepancy distributions of model-generated and human texts as well as the impact of passage length on detection. Figure~\ref{fig:perturbation-discrepancy-neo} shows the perturbation discrepancy distributions for model-generated and human texts across four data distributions, using GPT-Neo-2.7B to generate samples. A threshold of slightly below 0.1 separates human and model texts across data distributions, which is important for practical scenarios in which a passage may be analyzed without knowing its domain a priori. Finally, Figure~\ref{fig:auroc-vs-len} shows an analysis of {\name}'s performance as a function of passage length. We bin the paired human- and model-generated sequences by their average length into three bins of equal size (bottom/middle/top third), and plot the AUROC within each bin. The relationship between detection performance and passage length generally depends on the dataset and model (or tokenizer). For very long sequences, {\name} may see reduced performance because our implementation of {\name} applies all T5 mask-filling perturbations at once, and T5 may fail to track many mask tokens at once. By applying perturbations in multiple sequential rounds of smaller numbers of masks, this effect may be mitigated.

\section{Discussion}

As large language models continue to improve, they will become increasingly attractive tools for replacing human writers in a variety of contexts, such as education, journalism, and art. While legitimate uses of language model technologies exist in all of these settings, teachers, readers, and consumers are likely to demand tools for verifying the human origin of certain content with high educational, societal, or artistic significance, particularly when factuality (and not just fluency) is crucial.

In light of these elevated stakes and the regular emergence of new large language models, we study the \textit{zero-shot machine-generated text detection} problem, in which we use only the raw log probabilities computed by a generative model to determine if a candidate passage was sampled from it. We identify a property of the log probability function computed by a wide variety of large language models, showing that a tractable approximation to the trace of the Hessian of the model's log probability function provides a useful signal for detecting model samples. Our experiments find that this signal is more discriminative than existing zero-shot detection methods and is competitive with bespoke detection models trained with millions of model samples.

\noindent \textbf{{\name} and Watermarking.}
One interpretation of the perturbation function is producing \textit{semantically similar rephrasings of the original passage}. If these rephrasings are systematically lower-probability than the original passage, the model is exposing its bias toward the specific (and roughly arbitrary, by human standards) phrasing used. In other words, LLMs that do not perfectly imitate human writing essentially watermark themselves implicitly. Under this interpretation, efforts to \textit{manually} add watermarking biases to model outputs \citep{aaronson_2022,kirchenbauer2023watermark} may further improve the effectiveness of methods such as {\name}, even as LLMs continue to improve. 

\noindent \textbf{Limitations.}
One limitation of probability-based methods for zero-shot {\probfull} (like {\name}) is the white-box assumption that we can evaluate log probabilities of the model(s) in question. For models behind APIs that do provide probabilities (such as GPT-3), evaluating probabilities nonetheless costs money.
Another assumption of {\name} is access to a reasonable perturbation function. While in this work, we use off-the-shelf mask-filling models such as T5 and mT5 (for non-English languages), some domains may see reduced performance if existing mask-filling models do not well represent the space of meaningful rephrases, reducing the quality of the curvature estimate. While {\name} provides the best available detection performance for PubMedQA, its drop in performance compared to other datasets may be a result of lower quality perturbations. Finally, {\name} is more compute-intensive than other methods for detection, as it requires sampling and scoring the set of perturbations for each candidate passage, rather than just the candidate passage; a better tuned perturbation function or more efficient curvature approximation may help mitigate these costs.

\noindent \textbf{Future Work.}
While the methods in this work make no assumptions about the models generating the samples, future work may explore how watermarking algorithms can be used in conjunction with detection algorithms like {\name} to further improve detection robustness as language models continually improve their reproductions of human text. Separately, the results in Section~\ref{sec:variants} suggest that extending {\name} to use ensembles of models for scoring, rather than a single model, may improve detection in the black box setting. Another topic that remains unexplored is the relationship between prompting and detection; that is, can a clever prompt successfully prevent a model's generations from being detected by existing methods? Finally, future work may explore whether the local log probability curvature property we identify is present for generative models in other domains, such as audio, video, or images. We hope that the present work serves as inspiration to future work developing effective, general-purpose methods for mitigating potential harms of machine-generated media.

\section*{Acknowledgements}
EM gratefully acknowledges funding from a Knight-Hennessy Graduate Fellowship. CF and CM are CIFAR Fellows. The Stanford Center for Research on Foundation Models (CRFM) provided part of the compute resources used for the experiments in this work.

\bibliography{main}
\bibliographystyle{icml2023}

\newpage
\appendix
\onecolumn
\section{Complete Results for Top-$p$ and Top-$k$ Decoding}
Tables~\ref{tab:main-results-topp} and~\ref{tab:main-results-topk} contain the complete results for XSum, SQuAD, and WritingPrompts for the five models considered in Table~\ref{tab:main-results}. On average, both top-$p$ and top-$k$ sampling seem to make the detection task easier. This result is perhaps intuitive, as both sampling methods strictly increase the average log likelihood of model generations under the model (as they truncate low-probability tokens, albeit with different heuristics). Therefore methods based on probability or rank of tokens should become more discriminative.

\addtolength{\tabcolsep}{-0.2em}
\begin{table*}[h]
    \centering
    \small
    \resizebox{\textwidth}{!}{%
    \begin{tabular}{lcccccc|cccccc|cccccc}
         & \multicolumn{6}{c}{\textbf{XSum}} & \multicolumn{6}{c}{\textbf{SQuAD}} & \multicolumn{6}{c}{\textbf{WritingPrompts}} \\
         \cmidrule(lr){2-7} \cmidrule(lr){8-13} \cmidrule(lr){14-19}
         \textbf{Method} & GPT-2 & OPT-2.7 & Neo-2.7 & GPT-J & NeoX & \textbf{Avg.} & GPT-2 & OPT-2.7 & Neo-2.7 & GPT-J & NeoX & \textbf{Avg.} & GPT-2 & OPT-2.7 & Neo-2.7 & GPT-J & NeoX & \textbf{Avg.} \\
         \midrule
         $\log p(x)$ & 0.93 & 0.93 & 0.94 & 0.91 & 0.87 & 0.92 & 0.96 & 0.94 & 0.91 & 0.87 & 0.79 & 0.89 & \phantom{*}0.99* & \phantom{*}0.98* & \phantom{*}0.98* & \phantom{*}0.97* & \phantom{*}0.97* & \textbf{0.98} \\
        Rank & 0.80 & 0.77 & 0.77 & 0.75 & 0.73 & 0.76 & 0.84 & 0.82 & 0.81 & 0.80 & 0.75 & 0.81 & 0.87 & 0.84 & 0.83 & 0.83 & 0.81 & 0.84 \\
        LogRank & \phantom{*}0.95* & \phantom{*}0.94* & \phantom{*}0.96* & \phantom{*}0.93* & \phantom{*}0.89* & \phantom{*}0.93* & \phantom{*}0.98* & \phantom{*}0.96* & \phantom{*}0.94* & \textbf{0.90} & \textbf{0.83} & \phantom{*}0.92* & \phantom{*}0.99* & \phantom{*}0.98* & \phantom{*}0.98* & \textbf{0.98} & \textbf{0.98} & \textbf{0.98} \\
        Entropy & 0.55 & 0.46 & 0.53 & 0.54 & 0.58 & 0.53 & 0.53 & 0.50 & 0.55 & 0.56 & 0.57 & 0.54 & 0.32 & 0.37 & 0.28 & 0.32 & 0.32 & 0.32 \\
        {\name} & \textbf{0.99} & \textbf{0.98} & \textbf{1.00} & \textbf{0.98} & \textbf{0.97} & \textbf{0.98} & \textbf{0.99} & \textbf{0.98} & \textbf{0.98} & \textbf{0.90} & \phantom{*}0.82* & \textbf{0.94} & \textbf{1.00} & \textbf{0.99} & \textbf{0.99} & \phantom{*}0.97* & 0.93 & \textbf{0.98} \\
        \midrule
        Diff & 0.04 & 0.04 & 0.04 & 0.05 & 0.08 & 0.05 & 0.01 & 0.02 & 0.04 & 0.00 & -0.01 & 0.02 & 0.01 & 0.01 & 0.01 & -0.01 & -0.05 & 0.00 \\
         \bottomrule
    \end{tabular}}
    \caption{Nucleus (top-$p$) sampling evaluation with $p=0.96$. AUROC for detecting samples from the given model on the given dataset for {\name} and four previously proposed criteria. Nucleus sampling generally makes detection easier for all methods, but {\name} still provides the highest average AUROC. For WritingPrompts, however, the LogRank baseline performs as well as {\name}.}
    \label{tab:main-results-topp}
\end{table*}
\addtolength{\tabcolsep}{0.2em}

\addtolength{\tabcolsep}{-0.2em}
\begin{table*}[h]
    \centering
    \small
    \resizebox{\textwidth}{!}{%
    \begin{tabular}{lcccccc|cccccc|cccccc}
        \toprule
         & \multicolumn{6}{c}{\textbf{XSum}} & \multicolumn{6}{c}{\textbf{SQuAD}} & \multicolumn{6}{c}{\textbf{WritingPrompts}} \\
         \cmidrule(lr){2-7} \cmidrule(lr){8-13} \cmidrule(lr){14-19}
         \textbf{Method} & GPT-2 & OPT-2.7 & Neo-2.7 & GPT-J & NeoX & \textbf{Avg.} & GPT-2 & OPT-2.7 & Neo-2.7 & GPT-J & NeoX & \textbf{Avg.} & GPT-2 & OPT-2.7 & Neo-2.7 & GPT-J & NeoX & \textbf{Avg.} \\
         \midrule
         $\log p(x)$ & 0.89 & 0.89 & 0.89 & 0.84 & 0.81 & 0.87 & 0.93 & 0.90 & 0.88 & 0.82 & 0.74 & 0.85 & 0.97 & 0.95 & 0.97 & 0.96 & \phantom{*}0.95* & 0.96 \\
        Rank & 0.79 & 0.77 & 0.77 & 0.75 & 0.73 & 0.76 & 0.84 & 0.82 & 0.80 & 0.80 & 0.75 & 0.80 & 0.87 & 0.84 & 0.83 & 0.82 & 0.81 & 0.83 \\
        LogRank & \phantom{*}0.92* & \phantom{*}0.91* & \phantom{*}0.93* & \phantom{*}0.89* & \phantom{*}0.85* & \phantom{*}0.90* & \phantom{*}0.96* & \phantom{*}0.94* & \phantom{*}0.92* & \phantom{*}0.87* & \phantom{*}0.79* & \phantom{*}0.90* & \phantom{*}0.98* & \phantom{*}0.97* & \phantom{*}0.98* & \textbf{0.97} & \textbf{0.96} & \textbf{0.97} \\
        Entropy & 0.58 & 0.49 & 0.55 & 0.56 & 0.59 & 0.55 & 0.55 & 0.52 & 0.56 & 0.56 & 0.58 & 0.56 & 0.35 & 0.41 & 0.30 & 0.34 & 0.37 & 0.35 \\
        {\name} & \textbf{0.99} & \textbf{0.97} & \textbf{0.99} & \textbf{0.96} & \textbf{0.96} & \textbf{0.98} & \textbf{0.99} & \textbf{0.98} & \textbf{0.98} & \textbf{0.89} & \textbf{0.80} & \textbf{0.93} & \textbf{0.99} & \textbf{0.98} & \textbf{0.99} & \textbf{0.97} & 0.93 & \textbf{0.97} \\
        \midrule
        Diff & 0.07 & 0.06 & 0.06 & 0.07 & 0.11 & 0.08 & 0.03 & 0.04 & 0.06 & 0.02 & 0.01 & 0.03 & 0.01 & 0.01 & 0.01 & 0.00 & -0.03 & 0.00 \\
         \bottomrule
    \end{tabular}}
    \caption{Top-$k$ sampling evaluation with $k=40$. {\name} generally provides the most accurate performance (highest AUROC), although the gap is narrowed comparing to direct sampling, presumably because top-$k$ generations are more generic.}
    \label{tab:main-results-topk}
\end{table*}
\addtolength{\tabcolsep}{0.2em}

\end{document}